\documentclass[conference]{IEEEtran}
\IEEEoverridecommandlockouts
\usepackage{cite}
\usepackage{amsmath,amssymb,amsfonts}
\usepackage{algorithmic}
\usepackage{graphicx}
\usepackage{textcomp}
\usepackage{xcolor}
\usepackage{wrapfig}
\usepackage{booktabs}   
\usepackage{hyperref}
\usepackage{multirow}
\usepackage{makecell}
\usepackage{placeins}
\usepackage{stfloats}
\usepackage{subcaption}

\usepackage{tabularx}

\usepackage{algorithm}      
\DeclareUnicodeCharacter{00A0}{~}
\DeclareUnicodeCharacter{00B5}{\ensuremath{\mu}}
\DeclareUnicodeCharacter{00D7}{\ensuremath{\times}}
\DeclareUnicodeCharacter{03BC}{\ensuremath{\mu}}
\DeclareUnicodeCharacter{2013}{--}
\DeclareUnicodeCharacter{2019}{'}

\def\BibTeX{{\rm B\kern-.05em{\sc i\kern-.025em b}\kern-.08em
    T\kern-.1667em\lower.7ex\hbox{E}\kern-.125emX}}
\begin{document}

\title{Ranking Before Serving: Low-Latency LLM Serving via Pairwise Learning-to-Rank\\
}

\author{
Yiheng Tao$^{1}$, 
Yihe Zhang$^{1}$, 
Matthew Dearing$^{1,2}$, 
Xin Wang$^{1}$, 
Yuping Fan$^{2}$, 
Michael E. Papka$^{1,2}$, 
Zhiling Lan$^{1,2}$ \\[4pt]
$^{1}$University of Illinois Chicago, Chicago, IL, USA \\
$^{2}$Argonne National Laboratory, Lemont, IL, USA \\[4pt]
\{ytao28, yihe6, mdear2, xwang823, papka, zlan\}@uic.edu, 
\{fany, papka\}@anl.gov
}

\maketitle

\begin{abstract}
Efficient scheduling of large language model (LLM) inference tasks is critical for achieving low latency and high throughput, a challenge that is becoming increasingly acute with the rise of reasoning-capable LLMs whose generation lengths are highly variable.
Traditional strategies like First Come, First-Serve (FCFS) often suffer from Head-of-Line (HOL) blocking, where long-running tasks delay shorter ones queued behind them. 
In this paper, we introduce PARS, a prompt-aware LLM task scheduler that mitigates HOL blocking by approximating shortest-job-first (SJF) scheduling through pairwise ranking with a margin ranking loss.
PARS effectively predicts response-length–based task ordering directly from prompts, thereby optimizing scheduling decisions with minimal overhead. In addition, it integrates seamlessly with vLLM, a state-of-the-art LLM serving system, for the research community. Extensive experiments across multiple LLM models and real-world inference use cases (i.e., chat, math, and code generation) demonstrate that PARS significantly reduces latency by up to 15.7× compared to the vLLM default scheduler.  Cross-model evaluations demonstrate that our design generalizes effectively, allowing effective scheduling across diverse LLMs without requiring model-specific retraining.
\end{abstract}

\section{Introduction}

In recent years, large language models (LLMs) have emerged as core engines for artificial intelligence applications, demonstrating remarkable capabilities in a wide range of tasks, including question answering, code generation, and text classification. Deployments with these generative models, particularly in interactive scenarios, are expected to deliver real-time responses under high user traffic and heavy request loads.

Recently a new generation of reasoning LLMs has emerged. Unlike previous instruction-following models that focus on delivering concise, direct responses, reasoning-oriented models, such as DeepSeek R1\cite{deepseekai2025deepseekr1incentivizingreasoningcapability}, OpenAI’s GPT-o1/o3\cite{openai2024openaio1card, O3guan2025deliberativealignmentreasoningenables}, and Phi-4-reasoning\cite{Phi-4-reasoning}
are designed to engage in a multi-step Chain of Thought (CoT)\cite{Chain-of-Thought} reasoning that generates intermediate thoughts, self-reflections, or justifications before producing a final response. 
While this evolution boosts problem-solving capabilities, it also introduces \emph{new challenges for model serving}, particularly due to increased and highly variable response lengths.

Due to the auto-regressive nature of LLM generation, where outputs are produced token-by-token, the inference time is primarily determined by the response length \cite{fuEfficientLLMScheduling2024, DistServe, PO}. 
This effect becomes more pronounced for reasoning LLMs, where output lengths can vary drastically depending on task complexity. For example, a simple factual query may require only a brief response, while a complex reasoning task, such as mathematical derivation or a multi-step logical procedure, can result in long outputs spanning thousands of tokens. As shown in Table~\ref{tab:model_token_lengths}, output lengths can vary by several orders of magnitude across different requests and LLM models.  This wide variability leads to \emph{highly unpredictable and non-uniform inference times across requests and LLMs}. 
\begin{table}[t]
  \centering
  \caption{Output length differences among LLMs for two example prompts:
    \emph{(Q1) How many \texttt{r} in ``Strawberry?''}
    \emph{(Q2) How many prime numbers are there $<\!10000$?}}
  \label{tab:model_token_lengths}
  \footnotesize
  \begin{tabular}{@{}lcc@{}}
    \toprule
    \textbf{Model} & \textbf{Reasoning} & \textbf{No. of Output Tokens} \\
    \midrule
    GPT\textendash4 \cite{GPT4}        & $\times$     & 14 (Q1), 15 (Q2) \\
    Llama 3.1 \cite{llama3}            & $\times$     & 1 (Q1),  5 (Q2)  \\
    GPT o3 \cite{O3guan2025deliberativealignmentreasoningenables} & $\checkmark$ & 3091 (Q1), 7285 (Q2) \\
    DS\textendash R1 \cite{deepseekai2025deepseekr1incentivizingreasoningcapability} & $\checkmark$ & 2751 (Q1), 8077 (Q2) \\
    \bottomrule
  \end{tabular}
\end{table}

\begin{figure}
    \centering
    \includegraphics[width=1\linewidth]{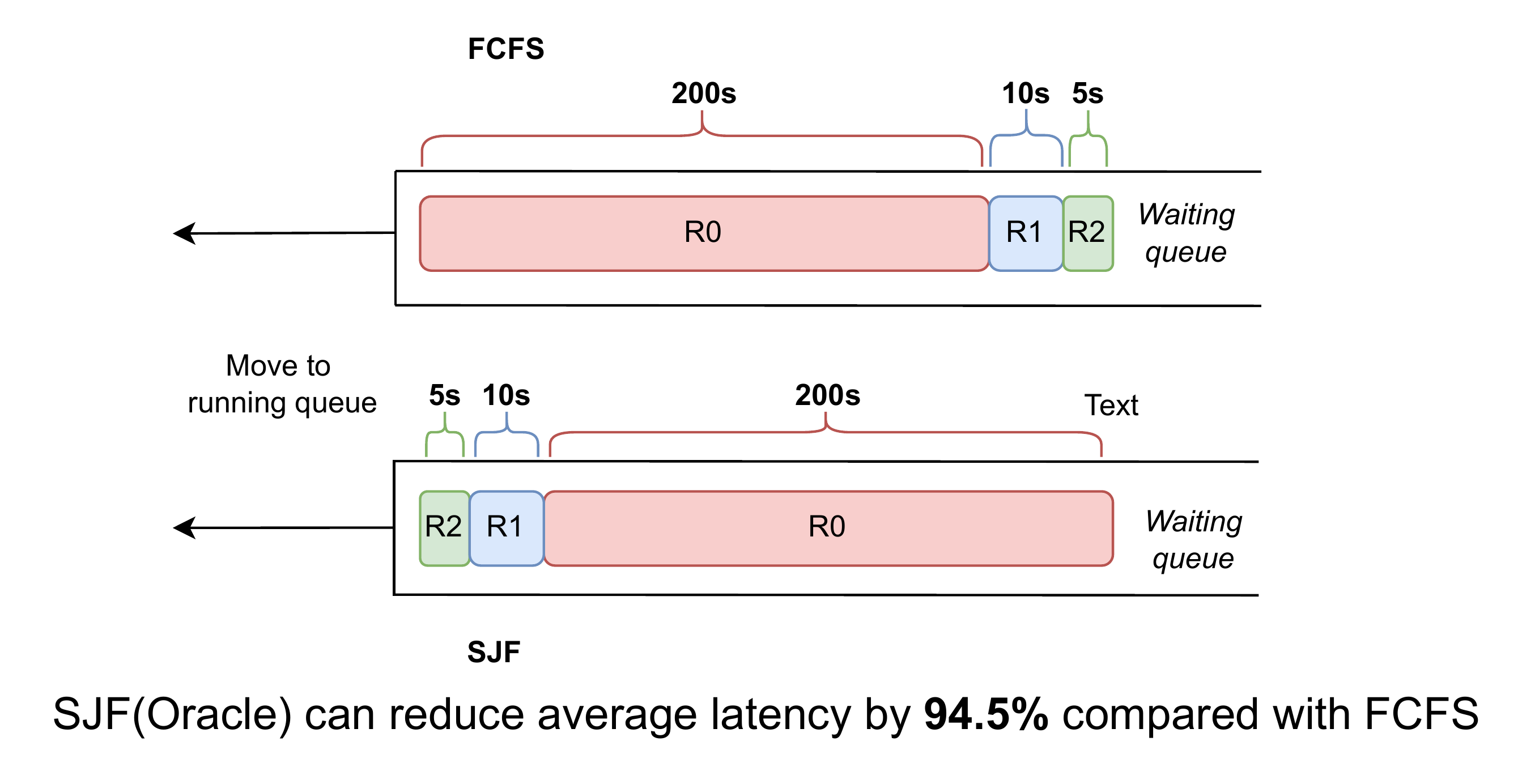}
    \caption{An example showing how Shortest-Job-First (SJF) scheduling can mitigate the head-of-line (HOL) blocking issue.}
    \label{fig:impact_of_sjf}
\end{figure}

Most state-of-the-art LLM serving systems, such as vLLM \cite{vLLM} and Orca \cite{orca}, currently adopt the First Come, First Serve (FCFS) scheduling policy, which processes requests in order of arrival. While simple and fair, FCFS suffers from the well-known issue of Head-Of-Line (HOL) blocking \cite{headofline1, headofline2}, where long-running requests block shorter ones behind them in the queue. To mitigate this, an alternative is the Shortest-Job-First (SJF) policy, which prioritizes requests with shorter expected execution times. Figure~\ref{fig:impact_of_sjf} depicts an illustrative example in which three inference tasks with response lengths of 200, 10, and 5 tokens, respectively, are waiting in the queue. Assume serving one token takes 1.0 second. Under FCFS, the average latency is 21.67 seconds per token. By reordering the serving sequence, SJF yields an average latency of 1.19 seconds per token (i.e., an 94.5\% reduction). This example demonstrates how intelligent scheduling can dramatically lower latency in heterogeneous workloads. As reasoning LLMs become more prevalent, we expect greater variability in inference tasks, highlighting the growing importance of scheduling strategies such as SJF for LLM serving systems.

However, effective SJF use requires knowing inference tasks’ response lengths in advance. While several studies have investigated approaches to predict the output length of each request prior to execution in order to approximate SJF\cite{PO, qiuEfficientInteractiveLLM2024a, qiuPowerawareDeepLearning}, 
\emph{accurately predicting output length for LLMs remains challenging} \cite{PO, qiuEfficientInteractiveLLM2024a, qiuPowerawareDeepLearning}.
The autoregressive and stochastic nature of LLM generation leads to varying response lengths for a given prompt across different runs. This challenge is further amplified in \emph{reasoning LLMs}, where multi-step reasoning traces introduce even greater variability and unpredictability in output lengths. This inherent variability makes it extremely challenging to predict exact output lengths ahead of time. An alternative is to bypass length prediction and instead learn a ranking over tasks \cite{fuEfficientLLMScheduling2024}. However, this approach is prone to noise. Because response lengths for similar prompts often fluctuate within a narrow range, task A may appear longer than task B in one run and shorter in another, for example. As a result, learning from such unstable rankings introduces inconsistent training signals and limits the effectiveness of scheduling decisions.

To bridge the gap, we introduce \textbf{PARS} (Prompt-Aware Ranking Scheduler), 
a lightweight and deployable LLM inference serving framework designed to mitigate HOL blocking in modern LLM workloads. 
At its core is \emph{a predictor} trained using a pairwise learning-to-rank approach with margin ranking loss.
Margin ranking loss is a pair-based loss that drives a model to produce a desired ranking or separation margin between pairs \cite{marginRanking1, marginRanking2, marginRanking3}. 
Our predictor is designed to focus on meaningful comparisons while disregarding noisy or low-impact data. This enables it to assign each incoming prompt a score reflecting its relative expected cost, which is sufficient for effective scheduling decisions.
At runtime, \emph{a predictor-guided scheduler} leverages the ranking scores to dynamically prioritize shorter tasks and approximate SJF scheduling,  
thereby reducing HOL blocking, lowering latency, and boosting throughput
with minimal overhead.

In addition, we implement PARS with vLLM and extensively evaluate it across multiple LLM models and real-world inference use cases.
Overall, our key contributions are:

\begin{itemize}
    \item We present a prompt-aware predictor that intelligently uses a pairwise learning-to-rank method with margin ranking loss to predict the output length of LLM inference tasks(\S\ref{section:predictor}). 
    \item We develop a predictor-guided scheduler that approximates SJF scheduling, explicitly accounting for highly variable generation lengths and full reasoning traces common in modern reasoning LLMs \footnote{For reasoning  LLMs, we explicitly include the full reasoning trace, including intermediate steps, thoughts, or justifications, as part of the response length, allowing our method to make effective scheduling decisions in complex multi-step generation scenarios.}. (\S\ref{section:scheduler}).
    \item We integrate our design into the state-of-the-art serving system vLLM as open-source software, making it immediately usable in both research and real-world deployment settings (\S\ref{section:imp}).
    \item We conduct extensive evaluations across multiple real-world datasets and LLM architectures, demonstrating that PARS consistently outperforms FCFS and existing SJF methods while exhibiting strong robustness under cross-model generalization (\S\ref{section:evaluation}).
\end{itemize}

\section{Background and Related Work}

\textbf{LLM inference} can be divided into two stages: \emph{prefilling}, where the model processes the input prompt in a single forward pass, and \emph{decoding}, where the model generates output tokens one at a time in an auto-regressive manner \cite{Survey1,Survey2,Survey3}. Prior work has shown that the decoding phase dominates the overall inference time, often accounting for the majority of latency in typical requests \cite{fuEfficientLLMScheduling2024,DistServe,PO}. As a result, predicting LLM generation length has become a practical and effective proxy for estimating task execution time in LLM serving systems.

In LLM inference systems, \emph{batching} strategies are crucial for optimizing computational efficiency and resource utilization\cite{orca,survey_book,Survey1,Survey2,Survey3}.

Static batching groups incoming requests into fixed-size batches that are executed together. While effective in offline or steady-load scenarios, it can introduce latency as requests must wait until the current batch is filled (or reaches a maximum waiting time) and the previous batch has completed. Consequently, static batching is less suitable for real-time applications requiring immediate responses.

Continuous batching (initially proposed by Orca \cite{orca}, also implemented in vLLM \cite{vLLM}) dynamically maintains the batch size during execution by substituting completed tasks with new incoming requests. In this mode, batches are updated at the iteration level: instead of waiting for an entire batch to complete, completed requests are immediately replaced with new ones from the queue.

Combining batching and scheduling strategies (such as approximating Shortest-Job-First) grants fine-grained control over which requests enter the batch and when, effectively resolving Head-of-Line (HOL) blocking and optimizing resource utilization. Our work supports both static and continuous batching.

\textbf{Learning to Rank (LTR)} \cite{Learningtorank} is a machine learning method designed to solve ranking problems commonly encountered in search engines \cite{Learningtorank}, recommendation systems \cite{LTR2}, and other information retrieval tasks \cite{LTR3,LTR4}. In LTR, models are trained on labeled data to effectively rank documents in order of relevance to a given query. LTR methods generally fall into three categories: (i) \emph{Pointwise} ranking treats ranking as a regression \cite{pointwise1} or classification \cite{pointwise2,pointwise3} task, predicting a relevance score or label for each item independently; (ii) \emph{Pairwise} ranking \cite{pairwise1,pairwise2, pairwise3,pairwise4,pairwise5,pairwise6} treats ranking as the task of learning relative preferences between document pairs; and (iii) \emph{Listwise} ranking \cite{listwise1,listwise2,listwise3,listwise4,listwise5} treats the entire document list as one instance, aiming to directly optimize global ranking performance.

Researchers have explored ranking-based approaches to predict LLM response lengths and to improve the efficiency of request scheduling. Zheng et al. \cite{PO} proposed a Perception Only (PO) approach that leverages the LLM's ability to predict its response length before generation. While intuitive, this method incurs high prediction overhead. To reduce this cost, later approaches employed lightweight proxy models, such as BERT-base, DistilBERT, and OPT, to estimate output lengths. Systems like S³ \cite{S$^3$}, TetriInfer \cite{TetriServe}, and DynamoLLM \cite{DynamoLLM} feature classification-based models to categorize requests into length buckets, while Qiu et al. \cite{qiuEfficientInteractiveLLM2024a, qiuPowerawareDeepLearning} formulated length prediction as a regression problem optimized using L1 loss. These methods follow a pointwise LTR approach, where the predictor model learns to assign a scalar length or label to each request. 

Rather than predicting specific output lengths, Fu et al. \cite{fuEfficientLLMScheduling2024} focused on the relative ordering of requests by training a predictor that optimizes the overall ranking accuracy of a prompt list, which is a listwise approach in the LTR taxonomy. Additionally, they introduced Kendall’s Tau as a metric to quantify ranking accuracy and demonstrated that higher Tau scores correlate with better scheduling performance under approximate SJF policies. 

These existing studies have several limitations: (1) they treat all ranking signals equally, resulting in noise and reduced scheduling accuracy when learning from prompts with similar output lengths; (2) they struggle when outputs contain multi-step reasoning traces from reasoning LLMs; and (3) they lack cross-model generalization, requiring expensive retraining for each new deployment.
The PARS framework addresses \emph{these limitations} by focusing on informative prompt pairs, handling multi-step reasoning traces, and generalizing across models without requiring retraining.

\begin{figure*}[t]
    \centering
    \includegraphics[width=0.88\linewidth]{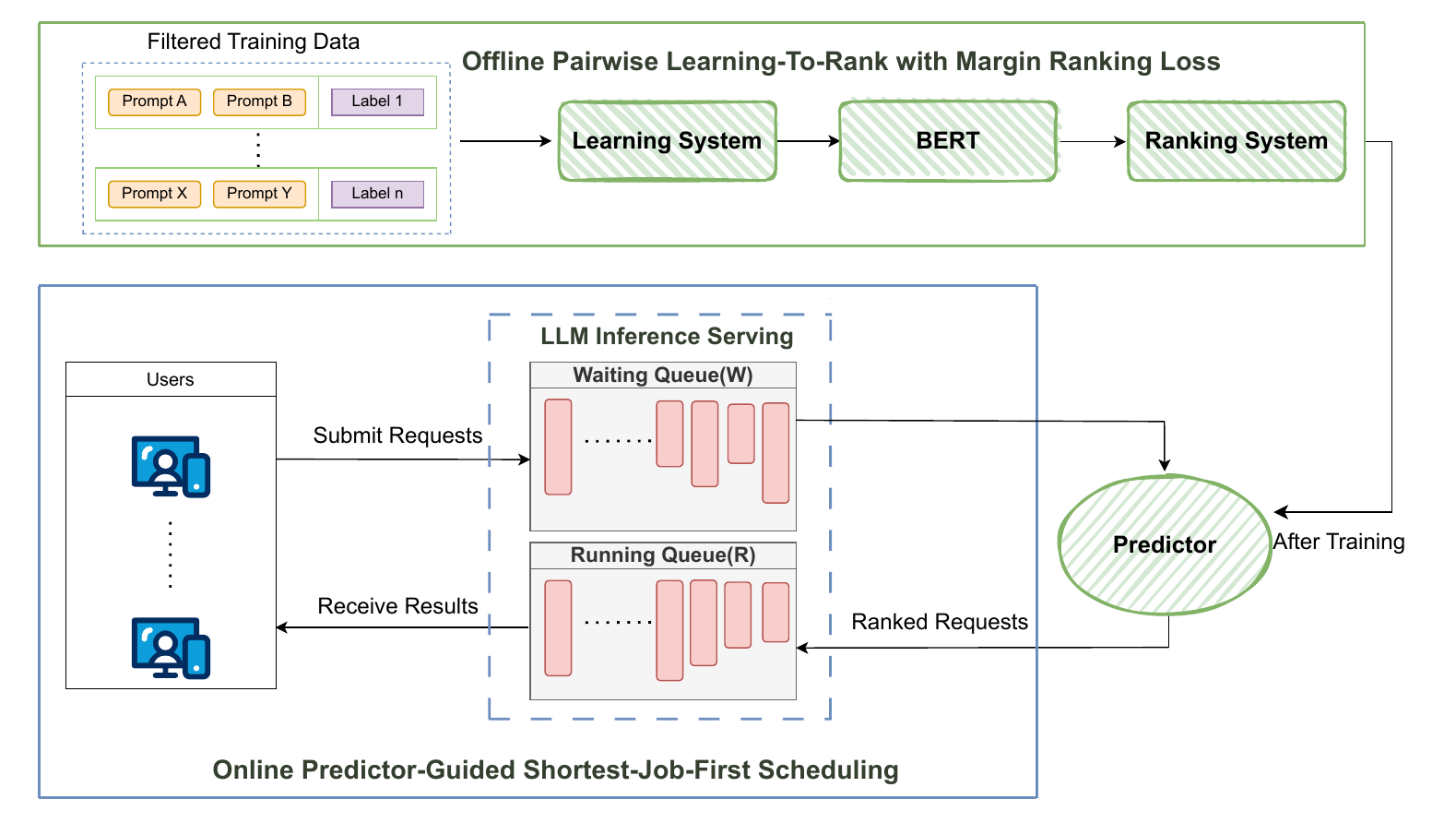}
    \caption{PARS overview: It consists of an offline training phase for constructing the predictor model and an online predictor-guided scheduler for deployment.}
    \label{fig:PARS}
\end{figure*}

\begin{table*}[t]
\centering
\caption{Pairwise ranking exhibits greater stability against stochastic LLM output lengths compared to pointwise and listwise ranking methods}

\label{tab:ltr_variance_case}
\footnotesize
\setlength{\tabcolsep}{4pt}
\renewcommand{\arraystretch}{1.15}

\begin{tabularx}{\textwidth}{l| X X |X X}
\toprule
\textbf{Method} &
\textbf{Input (Case 1)} &
\textbf{Learning Output (Case 1)} &
\textbf{Input (Case 2)} &
\textbf{Learning Output (Case 2)} \\
\midrule

\makecell[l]{\textbf{Pointwise}\\(e.g., REG-L1)} &
\makecell[l]{P1: 25 tok\\P2: 23 tok\\P3: 100 tok} &
\makecell[l]{\(s(P1)\approx 25\)\\\(s(P2)\approx 23\)\\\(s(P3)\approx 100\)} &
\makecell[l]{P1: 23 tok\\P2: 26 tok\\P3: 80 tok} &
\makecell[l]{\(s(P1)\approx 23\)\\\((s(P2)\approx 26\)\\\((s(P3)\approx 80\)} \\
\midrule

\makecell[l]{\textbf{Listwise}\\(e.g., ListMLE)} &
\makecell[l]{\(P2 < P1 < P3\)} &
\makecell[l]{\(s(P2) < s(P1) < s(P3)\)} &
\makecell[l]{\(P1 < P2 < P3\)} &
\makecell[l]{\(s(P1) < s(P2) < s(P3)\)} \\
\midrule

\makecell[l]{\textbf{Pairwise}\\(e.g., Margin Loss)} &
\makecell[l]{\(P1 \ll P3\), \(P2 \ll P3\)} &
\makecell[l]{\(s(P3) \gg s(P1)\)\\\(s(P3) \gg s(P2)\)} &
\makecell[l]{\(P1 \ll P3\), \(P2 \ll P3\)} &
\makecell[l]{\(s(P3) \gg s(P1)\)\\\(s(P3) \gg s(P2)\)} \\

\bottomrule
\end{tabularx}

\vspace{0.4em}
\parbox{\textwidth}{\footnotesize\textit{Note:} 
"tok" = tokens. \(P_i\) denotes a prompt instance. 
\(s(P)\) is the model's scoring output used for scheduling. 
Case 1 and Case 2 represent two separate sampling runs of the same prompts, illustrating  stochasticity of LLM
generation.
}

\end{table*}

\section{Methodology}

Figure~\ref{fig:PARS} depicts our framework for PARS (Prompt-Aware Ranking Scheduler) which consists of two components: (1) a \emph{predictor} trained with pairwise ranking and margin ranking loss, and (2) a \emph{scheduler} that uses the predictor for online request ranking combined with a SJF scheduling approach. 
    
\subsection{Novel Feature}
Before delving into the details of the PAR's components, we first highlight its novel feature. 

As mentioned previously, while prior methods have advanced length prediction and scheduling, they share a critical limitation: they \emph{underestimate the stochasticity of LLM generation}. In practice, the same prompt can yield different output lengths across multiple runs due to randomness in token sampling or temperature settings. 
This variance introduces significant label noise to learning objectives that rely on absolute values (pointwise) or full-order rankings (listwise), potentially confusing the model during training.

A novel feature of our work is \emph{pairwise ranking}. As illustrated by the example in Table~\ref{tab:ltr_variance_case}, existing length prediction approaches, such as pointwise and listwise methods, often produce inconsistent results under stochasticity. In contrast, pairwise learning focuses only on high-gap prompt pairs (e.g., $P_3 \gg P_1$) whose relative ordering tends to be preserved even under sampling variation. By avoiding near-tie comparisons that are sensitive to minor fluctuations, our pairwise training offers greater robustness and stability during learning.

\subsection{Pairwise-Trained Predictor} \label{section:predictor}
\textbf{Pairwise with Margin Ranking Loss.} Our predictor is designed with the intuition that determining which of two prompts results in a longer inference response is sufficient to order all prompts within an LLM inference batch. In other words, making accurate \emph{pairwise decisions} leads to a reliable global ranking. Therefore, we focus the training objective on maximizing the pairwise comparison accuracy.

Specifically, the training dataset consists of \emph{paired prompts} \((A, B)\), with each pair annotated by a binary label \(y\) indicating which prompt is expected to generate a longer response by the LLM. The labeling scheme is defined as:
(i) \(y = 1\) $\rightarrow$ Prompt \(A\) is expected to yield a longer response than Prompt \(B\); or (ii) \(y = -1\) $\rightarrow$ Prompt \(A\) is expected to yield a shorter response than Prompt \(B\).

Due to the inherent stochasticity of LLM generation, a prompt’s response length can vary across runs. When two prompts have similar expected lengths, their actual outputs may fluctuate within a small range, making their relative ordering less meaningful and potentially detrimental for predictor learning. Therefore, we hypothesize that the predictor should focus on ranking only prompt pairs with substantial differences in expected response length and filter out pairs with negligible differences to ensure the ranking effectively supports scheduling decisions.

To verify this hypothesis, we experimented by randomly selecting 200 prompts from the real-world prompt dataset Alpaca \cite{alpaca} and running each prompt ten times independently through Llama 3.1 and DeepSeek R1. Under a commonly used decoding setup (temperature~=~0.7, top-\(p\)~=~0.9), we observed that the relative output length variance across repeated generations for the same prompt typically remained around 20\% for both models. This suggests that prompt pairs with small length differences may introduce noisy learning signals, as such differences often fall within the LLM’s natural variability and do not reflect meaningful distinctions in task complexity. Here, relative variance is defined as \[
\text{Relative Variance} = \left( \frac{\text{max\_output\_length}}{\text{min\_output\_length}} - 1 \right) \times 100\%
\]

Based on this evidence, we adopt a minimum threshold to focus our training on informative prompt pairs. In other words, for each training prompt pair \((A, B)\),  we require that their ground-truth response lengths differ by at least this relative amount, defined as:
\begin{equation}
\texttt{min\_length\_difference} = \frac{\left|L_A - L_B\right|}{\max(L_A, L_B)}
\end{equation}
Here, \(L_A\) and \(L_B\) denote the response lengths of prompts \(A\) and \(B\), respectively. A pair is included in training only if \(\texttt{min\_length\_difference} \geq \delta\), where \(\delta\) is a tunable threshold.

In practice, the appropriate choice of \(\delta\) depends on the target LLM and its sampling configuration, such as temperature and top-\(p\), which affect the inherent output variability. We use a standard decoding configuration consistent with prior experiments (temperature~=~0.7, top-\(p\)~=~0.9), we empirically set \(\delta = 0.2\) for all models, based on their observed output variance. We further evaluate the effectiveness of training data filtering in Section~\ref{section:training_data_filtering}.

We use \emph{margin ranking loss} to encode this comparison logic into model training \cite{marginRanking1, marginRanking2, marginRanking3}. This approach enforces that if one prompt outranks another, then its predicted score should exceed the other by at least a fixed margin. 
In our case, the loss is defined as 
$\mathcal{L}(s_A, s_B, y) = \max(0, -y \cdot (s_A - s_B) + \text{margin})$,
which encourages the model to assign a higher score to the prompt expected to yield a longer response. 
Here, $s_A$ and $s_B$ are the predicted scores for prompts $A$ and $B$, respectively, and $y \in \{1, -1\}$ indicates the ground-truth ordering ($y=1$ when prompt $A$ is expected to yield a longer response, and $y=-1$ otherwise). The hyperparameter \texttt{margin}, which specifies the minimum difference between the scores of correctly ranked pairs, is set to 1.0 in our experiments.

\textbf{Feature Extraction and Ranking.}  
To extract semantic features from each prompt, we implement the pre-trained BERT-base-uncased model \cite{bert}. Specifically, we take the embedding of the classify token (\texttt{[CLS]}) from the pooler output, which serves as a compact summary of the input prompt. This embedding captures the semantic content of the prompt and serves as its feature representation. We further explain and validate our choice of BERT as the backbone model in Section~\ref{section:backbone_choice}.

To convert this feature into a usable ranking signal, we apply a lightweight linear layer that maps the BERT output to a scalar score. This score reflects the likelihood that the prompt will generate a long response. Higher scores indicate longer expected outputs. During predictor-guided scheduling, we rank pending requests by their scores to approximate SJF.

\subsection{Predictor-Guided Scheduler} \label{section:scheduler}

Our predictor-guided scheduler leverages the trained predictor to rank inference tasks based on their estimated response lengths. By ordering tasks from shortest to longest, it enables an SJF scheduling strategy that prioritizes quicker tasks and improves overall throughput. This scheduler is implemented and integrated into the modern LLM serving system vLLM \cite{vLLM}, which supports two task queues:
\begin{itemize}
    \item \emph{Waiting queue (W)} contains all requests that have arrived but have not yet begun execution.
   \item \emph{Running queue (R)} stores all requests currently being processed by the inference engine.
\end{itemize}

The scheduling cycle proceeds as follows. For each request in the waiting queue $W$, the predictor computes a ranking score. The queue is then sorted based on these scores, giving higher priority to those tasks predicted to have shorter responses. The top-ranked requests, up to the current batch size limit, are selected for execution and moved into the running queue $R$.

To avoid the issue of indefinitely deferring long-running requests, we incorporate a \emph{starvation prevention} mechanism. Each request is time-stamped at arrival, and if its wait time exceeds a predefined threshold (default: 2 minutes), then its priority is boosted, ensuring fairness with minimal impact on short tasks. This threshold is configurable and can be adjusted to meet application-specific latency SLOs. The full scheduling procedure operates at the iteration level, dynamically updating rankings and queue composition during inference. 
This design is fully compatible with modern LLM inference engines, such as vLLM and Orca \cite{vLLM,orca}. 

\subsection{Implementation} \label{section:imp}
We incorporate the PARS framework within the state-of-the-art LLM serving system vLLM.
The implementation is written in Python and does not require modification to the underlying C/C++ and CUDA backends. The scheduler is integrated into the Python-based request handling and queue management logic of vLLM 0.8.5, with approximately 1,000 lines of code.

\section{Evaluation}\label{section:evaluation}

We conduct a series of experiments across multiple models, datasets, and scheduling scenarios.  Our experiments are structued to answer the following key questions:

\begin{itemize}
\item \textbf{Q1:} How accurate is our predictor in ranking inference tasks? (\S\ref{section:ranking_performance})
\item \textbf{Q2:} Why do we choose BERT as the backbone, and how would prediction accuracy change with other lightweight models? (\S\ref{section:backbone_choice})
\item \textbf{Q3:} What is the impact of training data filtering on prediction accuracy? (\S\ref{section:training_data_filtering})
\item \textbf{Q4:} How does PARS perform in terms of scheduling efficiency? (\S\ref{section:scheduling_performance})
\item \textbf{Q5:} How well does PARS generalize across different LLMs? (\S\ref{section:generality})
\item \textbf{Q6:} Does PARS preserve throughput and batching efficiency while reducing latency? (\S\ref{section:throughput})
\item \textbf{Q7:} What is the impact of the starvation prevention mechanism?(\S\ref{section:starvation})
\item \textbf{Q8:} What is the runtime overhead introduced by PARS?(\S\ref{section:overhead})
\end{itemize}

\textbf{Testbed.} All experiments are conducted on a server equipped with two NVIDIA A100 GPUs (40GB each) and an ARM-based Neoverse-N1 CPU.

\textbf{Models.} We evaluate our method on three representative LLMs:

\begin{itemize}
    \item \textit{Llama 3.1} \cite{llama3} An open-source LLM developed by Meta, used in various instruction-following applications. We use the 8B version in our experiments.
    \item \textit{GPT-4} \cite{GPT4}: A proprietary model developed by OpenAI, also used in various instruction-following applications.
    \item \textit{DeepSeek-R1} \cite{deepseekai2025deepseekr1incentivizingreasoningcapability}: A reasoning-oriented model that includes multi-step thinking traces in its output. We use the distilled 7B version based on Qwen\cite{qwen}. For this model, we include the full reasoning process as part of the generation length because these intermediate outputs contribute significantly to the user-experienced inference time in practical deployments.
\end{itemize}

We select these LLMs to cover both open-source and closed-source models, as well as reasoning-oriented models. Note that GPT-4 is a closed-source model not supported in vLLM, so we only report its prediction accuracy.

\textbf{Datasets.} We use four real-world prompt datasets:
\begin{itemize}
    \item \textit{Alpaca} \cite{alpaca}: A widely used instruction-tuning dataset containing diverse natural language prompts across a broad range of tasks.
    \item  \textit{LMSYS-Chat-1M} \cite{zheng2023lmsyschat1m}: A large-scale, multi-turn conversation dataset containing over one million real-world user interactions across multiple LLMs.
    \item  \textit{MATH-plus} \cite{MATHplus}: A math-focused dataset combining MetaMath\cite{Metamath}, Orca-Math\cite{Orcamath}, and additional GPT-4-augmented math problems.
    \item  \textit{Evolved Codealpaca} \cite{Wizardcoder}: A large-scale code generation dataset with diverse programming tasks (e.g., function writing, refactoring, bug fixing), designed to train and evaluate LLMs.
\end{itemize}

These datasets are chosen to cover diverse LLM use cases, including instruction-following, general conversation, mathematics problem solving, and code generation.

For each dataset, we randomly select 5,000 prompts for testing and 40,000 prompts are drawn from the remaining pool for training. Additionally, a subset of 2,000 prompts from the test set is reserved for bursty inference experiments. 

The PARS predictor and all baseline predictors are trained on the same training data for five epochs with a batch size of 128, using margin Ranking Loss and Adam optimizer with a fixed learning rate of 2e\text{-}5.

\textbf{Scheduling Policies.} We compare several scheduling methods in our experiments: 

\begin{itemize}
\item \textit{First Come, First Serve (FCFS)}: A widely used serving  policy that schedules inference requests in the order of their arrival. This serves as our baseline.
\item \textit{Pointwise SJF}: a shortest-job-first scheduling method using pointwise ranking regression with L1 loss \cite{qiuEfficientInteractiveLLM2024a,qiuPowerawareDeepLearning}.
\item \textit{Listwise SJF}: a shortest-job-first scheduling method using listwise ranking with ListMLE loss \cite{fuEfficientLLMScheduling2024}.
\item \textit{Oracle SJF}: An idealized shortest-job-first scheduler using ground-truth response lengths from prior runs to simulate perfect foresight, serving as the optimal bound in our experiments.
\item \textit{PARS}: our design integrating intelligent pairwise ranking and predictor-guided scheduling. In addition, we evaluate a variant of our framework called \textit{Cross-Model PARS}, in which the predictor is trained on GPT-4 model but applied to rank inference tasks for other LLMs. Cross-Model PARS is included to demonstrate the generalization capability of our predictor across models, enabling efficient scheduling even when the target models lack their own training data.
\end{itemize}

In our experiments, we report results using continuous batching with default vLLM batch size, as continuous batching is widely adopted in modern LLM serving systems for handling online requests \cite{vLLM}.

\textbf{Evaluation Metrics.} For evaluating predictor accuracy, we adopt the \textit{Kendall rank correlation coefficient} \cite{fuEfficientLLMScheduling2024}. 
Kendall’s Tau \cite{Kendall’sTau1} is a non-parametric rank correlation coefficient that measures the ordinal association between two ordered variables. Formally, it is defined as
$\tau_b = \frac{n_c - n_d}{\sqrt{(n_0 - n_1)(n_0 - n_2)}}$, 
where \(n\) is the total number of ranked items, \(n_c\) is the number of concordant pairs, \(n_d\) is the number of discordant pairs, $n_0=n(n-1)/2$ represents the total number of pairs, and \(n_1\) and \(n_2\) denote the number of tied pairs in the first and second data sets, respectively. The coefficient takes values in the range \([-1, 1]\), where \(-1\) indicates complete disagreement between the rankings, \(+1\) indicates perfect agreement, and \(0\) suggests no significant correlation in rank order.

To evaluate scheduling performance, we report the average and 90th‑percentile per‑token latency \cite{fuEfficientLLMScheduling2024, qiuEfficientInteractiveLLM2024a, qiuPowerawareDeepLearning}, defined as the end‑to‑end request latency divided by the output length. In the remainder of this paper, we refer to these metrics as \textit{average latency} and \textit{p90 latency}, respectively.

\subsection{Predictor Accuracy} \label{section:ranking_performance}

We compare the pairwise ranking strategy used in PARS against two alternative ranking approaches from prior work: Listwise Ranking (ListMLE) \cite{fuEfficientLLMScheduling2024} and Pointwise Ranking (regression with L1 loss) \cite{qiuPowerawareDeepLearning, qiuEfficientInteractiveLLM2024a}. The evaluation is conducted using four datasets (Alpaca \cite{alpaca, alpacagpt4}, LMSYS-Chat-1M \cite{zheng2023lmsyschat1m}, MATH-plus \cite{MATHplus} and Evolved Codealpaca \cite{Wizardcoder}) across three models (GPT-4, Llama 3.1, and DeepSeek-R1). For the reasoning-oriented model DeepSeek-R1, generation length is defined to include both the final response and intermediate reasoning steps.
These reasoning chains are part of the inference process, consuming computational resources and impacting scheduling performance, so they must be included in latency measurements.

Table \ref{Ranking_Performance} presents the $\tau_b$ values across various dataset and model combinations. 
First, the PARS predictor consistently achieves the highest ranking performance on all datasets, reaching a maximum accuracy of 0.96. On the LMSYS-Chat-1M dataset, for instance, pointwise SJF achieves only 0.09 accuracy on the R1 model, whereas PARS boosts it to 0.50, a more than 5.5× improvement. Second, prediction accuracy is generally higher with GPT-4 compared to Llama 3.1 and R1 on Alpaca, LMSYS-Chat-1M and Evol-CodeAlpaca, while on MATH-plus, R1 achieves slightly higher accuracy than GPT-4. Third, prediction accuracy varies across both datasets and models. It is highest on Alpaca (up to 0.96), moderate on LMSYS-Chat-1M (up to 0.72) and MATH-plus (up to 0.73), and lowest on Evol-CodeAlpaca (up to 0.62). At the dataset level, this reflects differences in task structure: simple instructional and conversational datasets are generally easier to predict, whereas math and code datasets tend to involve more complex prompts and less predictable outputs. At the model level, variation is attributed to differences in generation patterns across LLMs, which affect the predictability of output length. 
Finally, Cross-Model PARS demonstrates strong generalization ability. Even when trained solely on GPT-4,, it outperforms the pointwise approach and matches or exceeds the listwise approach trained individually on Llama and R1 in nearly all settings. This makes PARS ideal for deployment, as it can maintain robust performance across different LLM models without retraining.

\setlength{\tabcolsep}{2pt}      
\renewcommand{\arraystretch}{1.4}

\begin{table}[ht]
    \caption{Kendall’s Tau ($\tau_b$) comparison across datasets, LLMs, and ranking approaches. Cross-Model PARS is trained on GPT-4 and applied on Llama and R1.}
    \centering
    \resizebox{\columnwidth}{!}{
    \begin{tabular}{lccc|c}
    \toprule
    \textbf{Dataset} & \textbf{Listwise} & \textbf{Pointwise} & \textbf{PARS (Pairwise)} & \textbf{Cross-Model PARS}\\
    \midrule
    Alpaca (GPT-4) & 0.69 & 0.70 & \textbf{0.96}  & -\\
    Alpaca (Llama) & 0.67 & 0.64 & \textbf{0.75} & 0.65 \\
    Alpaca (R1)    & 0.50 & 0.30 & \textbf{0.61} & 0.50\\
    LMSYS-Chat-1M (GPT-4) &0.63  &0.33  & \textbf{0.72} & -\\
    LMSYS-Chat-1M (Llama) &0.54  & 0.37 & \textbf{0.65} & 0.62\\
    LMSYS-Chat-1M (R1) & 0.35 & 0.09 & \textbf{0.50} & 0.44\\
    MATH-plus (GPT-4) & 0.48 & 0.54 & \textbf{0.71} & - \\
    MATH-plus (Llama) & 0.52 & 0.54 & \textbf{0.65} & 0.54 \\
    MATH-plus (R1) & 0.63 & 0.55 & \textbf{0.73} & 0.53 \\
    Evol-CodeAlpaca (GPT-4) & 0.54 & 0.54 & \textbf{0.62} & - \\
    Evol-CodeAlpaca (Llama) & 0.55 & 0.55 & \textbf{0.61} & 0.55 \\
    Evol-CodeAlpaca (R1) & 0.39 & 0.34 & \textbf{0.46} & 0.35 \\
    \bottomrule
    \end{tabular}
    }
    \label{Ranking_Performance}
\end{table}

\subsection{Why We Use BERT} \label{section:backbone_choice}

In designing the PARS predictor, the choice of backbone model is guided by three key factors: (i) lightweight inference to avoid becoming a serving bottleneck, (ii) low training cost, and (iii) high prediction accuracy. In addition, the success of transfer learning in natural language processing (NLP), where a model is first pretrained on a data-rich task and then fine-tuned on a downstream task\cite{transferlearning}. Prior work also shows that pretrained Transformer models can achieve strong performance even when downstream training data is limited, by leveraging their pretrained representations~\cite{bert}, \cite{OPT}, \cite{T5}. These observations lead us to focus on \emph{lightweight pretrained Transformer backbones} for the scheduling predictor.

We compare three representative Transformer architectures with distinct designs: BERT (encoder-only)\cite{bert}, OPT (decoder-only)\cite{OPT}, and T5 (encoder-decoder)\cite{T5}. Table~\ref{tab:backbone_compare} summarizes the ranking performance ($\tau_b$ values) of these Transformer backbones across various datasets and LLMs under our pairwise training method. We observe that our method is consistently effective across all three backbones, demonstrating its architecture-agnostic nature. Among them, BERT achieves the highest prediction accuracy, motivating us to adopt BERT as the default backbone in our framework.

\setlength{\tabcolsep}{5pt}      
\renewcommand{\arraystretch}{1.4}

\begin{table}[ht]
    \caption{Kendall’s Tau ($\tau_b$) comparison across datasets, LLMs, and different Transformer backbones (using pairwise training).}
    \centering
    \footnotesize
    \begin{tabular}{lccc}
    \toprule
    \textbf{Dataset} & \textbf{T5} & \textbf{OPT} & \textbf{BERT} \\
    \midrule
    Alpaca (GPT-4) & 0.80 & 0.89 & \textbf{0.96} \\
    Alpaca (Llama) & 0.65 & 0.75 & \textbf{0.75} \\
    Alpaca (R1)    & 0.60 & 0.58 & \textbf{0.61} \\
    LMSYS-Chat-1M (GPT-4) & 0.70 & 0.70 & \textbf{0.72} \\
    LMSYS-Chat-1M (Llama) & 0.64 & 0.64 & \textbf{0.65} \\
    LMSYS-Chat-1M (R1) & 0.41 & 0.37 & \textbf{0.50} \\
    MATH-plus (GPT-4) & 0.68 & 0.70 & \textbf{0.71} \\
    MATH-plus (Llama) & 0.62 & 0.65 & \textbf{0.65} \\
    MATH-plus (R1) & 0.70 & 0.72 & \textbf{0.73} \\
    Evol-CodeAlpaca (GPT-4) & 0.62 & 0.61 & \textbf{0.62} \\
    Evol-CodeAlpaca (Llama) & 0.59 & 0.60 & \textbf{0.61} \\
    Evol-CodeAlpaca (R1) & 0.45 & 0.46 & \textbf{0.46} \\
    \bottomrule
    \end{tabular}
    \label{tab:backbone_compare}
\end{table}

\subsection{Impact of Training Data Filtering} \label{section:training_data_filtering}
Table~\ref{tab:min_length_difference} presents the prediction accuracy with and without applying \texttt{min\_length\_difference} filtering during training. Even without this filtering, our pairwise learning approach  achieves substantial gains over other learning-to-rank methods. By further filtering out low-impact training pairs, we improve the model’s focus on meaningful ranking signals, leading to consistent gains in prediction accuracy across datasets and models.
\setlength{\tabcolsep}{5pt}      
\renewcommand{\arraystretch}{1.4} 

\begin{table}[ht]
    \caption{Kendall’s Tau ($\tau_b$) comparison with and without \texttt{min\_length\_difference} filtering.}
    \centering
    \begin{tabular}{lcc}
    \toprule
    \textbf{Dataset} & \textbf{Without Filtering} & \textbf{With Filtering} \\
    \midrule
    Alpaca (GPT-4) & 0.93 & \textbf{0.96} \\
    Alpaca (Llama) & 0.71 & \textbf{0.75} \\
    Alpaca (R1)    & 0.57 & \textbf{0.61} \\
    LMSYS-Chat-1M (GPT-4) & 0.68 & \textbf{0.72} \\
    LMSYS-Chat-1M (Llama) & 0.62 & \textbf{0.65} \\
    LMSYS-Chat-1M (R1) & 0.46 & \textbf{0.50} \\
    MATH-plus (GPT-4) & 0.66  & \textbf{0.71} \\
    MATH-plus (Llama) & 0.52  & \textbf{0.65} \\
    MATH-plus (R1) & 0.69  & \textbf{0.73} \\
    Evol-CodeAlpaca (GPT-4) & 0.59 & \textbf{0.62} \\
    Evol-CodeAlpaca (Llama) & 0.59 & \textbf{0.61} \\
    Evol-CodeAlpaca (R1) &  0.43 & \textbf{0.46} \\
    \bottomrule
    \end{tabular}
    
    \label{tab:min_length_difference}
\end{table}

\subsection{Scheduling Efficiency} \label{section:scheduling_performance}

We evaluate our proposed scheduling policy in vLLM using eight \emph{(Dataset, Model)} combinations under realistic workloads.

\begin{figure*}
    \centering
    \includegraphics[width=1\linewidth]{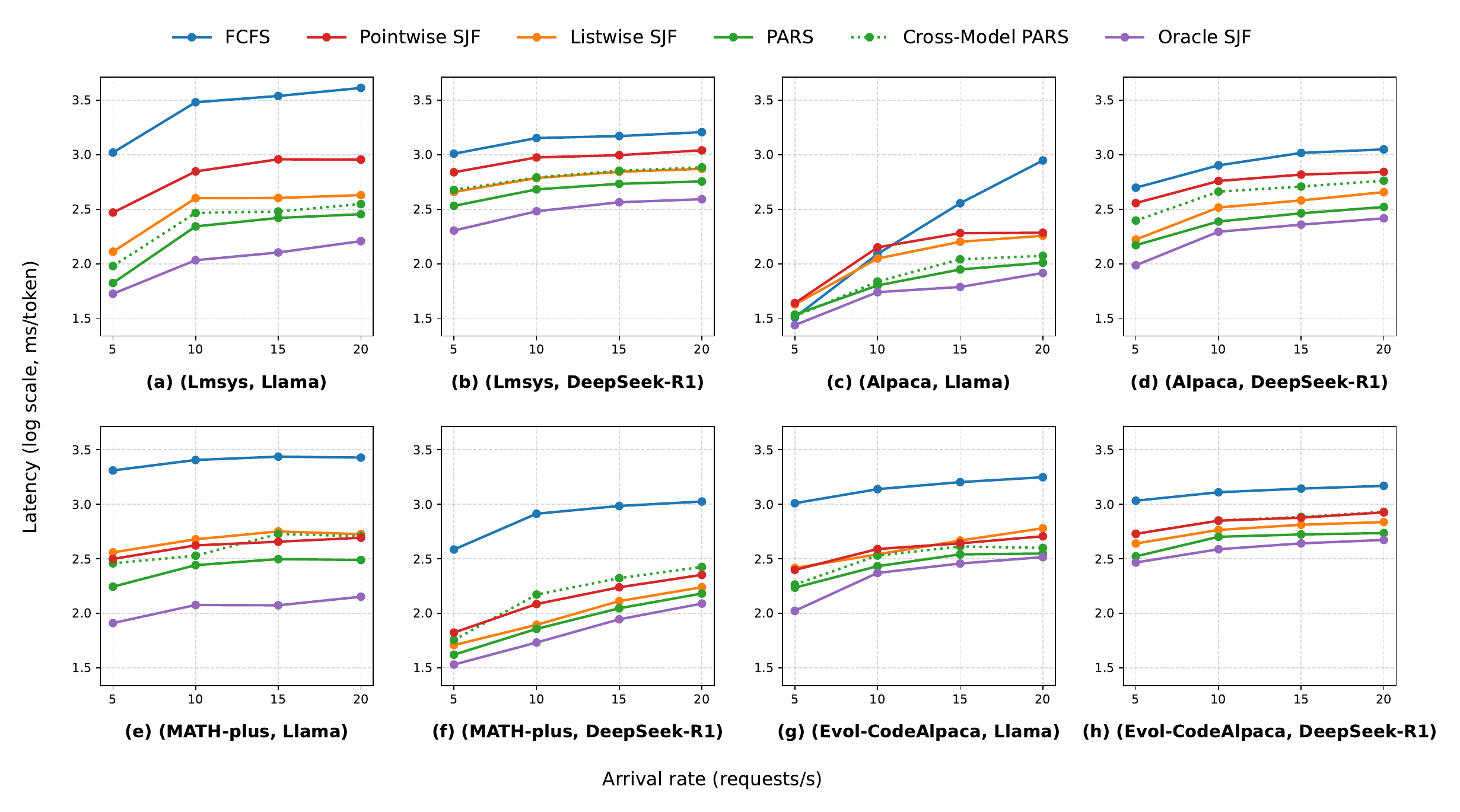}
    \caption{Average latency of different scheduling schemes with Llama and DeepSeek-R1 models on real workloads.}
    \label{fig:avg-latency}
\end{figure*}

Figure~\ref{fig:avg-latency} illustrates the average per-token latency across varying arrival rates for the eight \emph{(Dataset, Model)} combinations, comparing five scheduling strategies. In all cases, our scheduler consistently achieves the lowest latency among the three practical implementations, second only to the Oracle SJF baseline. Across all arrival rates, on both the Llama 3.1 and DeepSeek-R1 models, our approach incurs no more than 200~ms of additional average per-token latency compared to the Oracle baseline, achieving up to 15.7× improvement on the Llama model compared to the FCFS baseline. These results highlight the practicality and efficiency of our method in real-world LLM serving scenarios.

To assess performance under extreme load, we simulate a burst scenario with 2,000 simultaneous requests across all model–workload pairs (denoted as \textbf{burst mode}). Table~\ref{tab:burst_avg_latency_table} reports the average latency for this scenario. Our approach consistently outperforms FCFS and all approximate SJF methods and closely follows the Oracle baseline. Specifically, it achieves over 2× speedup versus the naive FCFS policy on the reasoning model and up to 7.7× on the Llama model, demonstrating robust performance under high concurrency.

Tail latency, especially the 90th percentile (p90), is critical to capture delays experienced by the slowest requests. 
Table~\ref{tab:consolidated_p90_latency} presents p90 latency results for the five policies. As expected, Oracle SJF consistently delivers the lowest tail latency. Across all configurations, PARS ranks first among all practical implementations, outperforming FCFS, Listwise, and Pointwise approaches. Under burst scenarios, it achieves over 2.5× speedup over FCFS on the reasoning LLM and up to 12.3× on the Llama model, all while maintaining tail latency closest to the Oracle baseline.

Overall, our PARS scheduling policy significantly reduces average latency 
and tail latency, especially under high-concurrency workloads.

\subsection{Cross-Model Generalization}
\label{section:generality}

To evaluate the generality of PARS across different LLM architectures, we trained a variant of PARS using training data generated from the GPT-4 model, referred to as Cross-Model PARS, with the same four prompt datasets described above. 
We then tested this GPT-4–trained PARS in the eight \emph{(Dataset, Model)} combinations, each using a different downstream LLM, to assess its cross-model scheduling performance, as shown in
Figure~\ref{fig:avg-latency} and Tables~\ref{tab:burst_avg_latency_table}-\ref{tab:consolidated_p90_latency}.
Despite being trained on a different LLM, Cross-Model PARS continues to deliver strong performance. It outperforms Pointwise SJF and matches or exceeds Listwise SJF in nearly all scenarios, especially on the Llama model. While a slight performance drop is observed on the reasoning LLM (DeepSeek-R1), this is mainly because GPT-4 and DeepSeek-R1 follow different generation patterns, leading to less accurate prediction and scheduling efficiency. Even so, Cross-Model PARS still achieves substantial latency reductions compared to FCFS, including at least 1.73× speedup in mean latency and 1.76× in p90 latency over FCFS under high-load (20 requests/s) conditions, and up to 11.6× (mean) and 8.8× (p90) on the Lmsys–Llama workload. These results demonstrate that PARS exhibits strong generality, outperforming the generalization capabilities of prior work~\cite{fuEfficientLLMScheduling2024, qiuEfficientInteractiveLLM2024a,S$^3$,PO}.

\begin{figure}[t]
    \centering
    \includegraphics[width=0.8\linewidth]{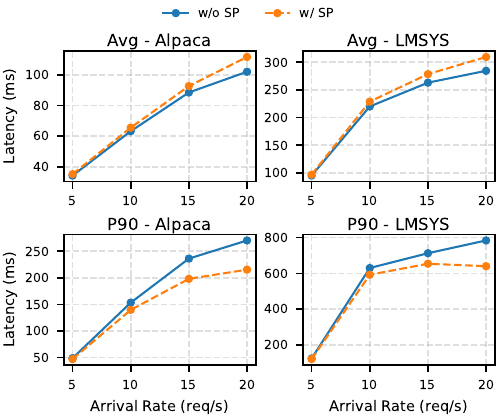}
    \caption{Latency (ms/token) with and without starvation prevention across arrival rates on two datasets under the Llama~3.1 model.}
    \label{fig:latency_starvation_comparison}
\end{figure}

\begin{table*}[t]
  \centering
  \caption{Average latency \textnormal{(ms/token)} under \textit{burst mode}. The best-performing approach highlighted in red, the second-best in blue, and the third-best in orange; the subsequent tables follow the same convention.}
  \label{tab:burst_avg_latency_table}
  \footnotesize
  \setlength{\tabcolsep}{6pt}
  \begin{tabular*}{\textwidth}{@{\extracolsep{\fill}}lcccccc@{}}
    \toprule
    \textbf{(Dataset, Model)} & \textbf{FCFS} & \textbf{Pointwise SJF} & \textbf{Listwise SJF} & \textbf{PARS} & \textbf{Cross-Model PARS} & \textbf{Oracle SJF} \\
    \midrule
    (Alpaca, DeepSeek-R1) & 372.03 & 283.39 & 187.41 & \textbf{\textcolor{blue}{148.97}} & \textbf{\textcolor{orange}{178.07}} & \textbf{\textcolor{red}{137.32}} \\
    (Alpaca, Llama)       & 962.53 & 152.30 & 157.07 & \textbf{\textcolor{blue}{124.92}} & \textbf{\textcolor{orange}{138.53}} & \textbf{\textcolor{red}{111.46}} \\
    (LMSYS-Chat-1M, DeepSeek-R1)  & 520.99 & 321.74 & 274.93 & \textbf{\textcolor{blue}{206.88}} & \textbf{\textcolor{orange}{255.90}} & \textbf{\textcolor{red}{153.24}} \\
    (LMSYS-Chat-1M, Llama)        & 1269.94 & 342.73 & 214.60 & \textbf{\textcolor{blue}{186.44}} & \textbf{\textcolor{orange}{187.86}} & \textbf{\textcolor{red}{143.19}} \\
    (MATH-plus, DeepSeek-R1) & 289.41 & 132.91 & \textbf{\textcolor{orange}{123.69}} & \textbf{\textcolor{blue}{115.65}} & 134.11 & \textbf{\textcolor{red}{107.10}} \\
    (MATH-plus, Llama)       & 298.35 & 167.76 & 165.87 & \textbf{\textcolor{blue}{154.25}} & \textbf{\textcolor{orange}{164.80}} & \textbf{\textcolor{red}{140.78}} \\
    (Evol-CodeAlpaca, DeepSeek-R1)  & 384.28 & 203.79 & \textcolor{orange}{200.65} & \textbf{\textcolor{blue}{171.69}} & 201.04 & \textbf{\textcolor{red}{154.52}} \\
    (Evol-CodeAlpaca, Llama)        & 384.18 & 184.75 & 182.40 & \textbf{\textcolor{blue}{171.62}} & \textbf{\textcolor{orange}{182.05}} & \textbf{\textcolor{red}{155.97}} \\
    \bottomrule
  \end{tabular*}
\end{table*}

\begin{table*}[t]
  \centering
  \caption{p90 latency \textnormal{(ms/token)} under \textit{burst mode}}
  \label{tab:burst_p90_latency_table}
  \footnotesize
  \setlength{\tabcolsep}{6pt}
  \begin{tabular*}{\textwidth}{@{\extracolsep{\fill}}lcccccc@{}}
    \toprule
    \textbf{(Dataset, Model)} & \textbf{FCFS} & \textbf{Pointwise SJF} & \textbf{Listwise SJF} & \textbf{PARS} & \textbf{Cross-Model PARS} & \textbf{Oracle SJF} \\
    \midrule
    (Alpaca, DeepSeek-R1) & 878.68  & 705.24 & 386.81 & \textbf{\textcolor{blue}{258.05}} & \textbf{\textcolor{orange}{333.35}} & \textbf{\textcolor{red}{229.76}} \\
    (Alpaca, Llama)       & 2225.68 & 230.57 & 215.46 & \textbf{\textcolor{blue}{180.03}} & \textbf{\textcolor{orange}{208.03}} & \textbf{\textcolor{red}{157.61}} \\
    (LMSYS-Chat-1M, DeepSeek-R1)  & 1091.11 & 666.79 & 585.48 & \textbf{\textcolor{blue}{422.57}} & \textbf{\textcolor{orange}{548.46}} & \textbf{\textcolor{red}{259.01}} \\
    (LMSYS-Chat-1M, Llama)        & 2307.54 & 553.46 & 306.06 & \textbf{\textcolor{blue}{261.00}} & \textbf{\textcolor{orange}{262.05}} & \textbf{\textcolor{red}{208.75}} \\
    (MATH-plus, DeepSeek-R1) & 645.43 & 231.73 & \textbf{\textcolor{orange}{208.37}} & \textbf{\textcolor{blue}{185.06}} & 232.14 & \textbf{\textcolor{red}{157.83}} \\
    (MATH-plus, Llama)       &  632.70 & 292.77 & 282.40 & \textbf{\textcolor{blue}{249.51}} & \textbf{\textcolor{orange}{281.86}} & \textbf{\textcolor{red}{206.64}} \\
    (Evol-CodeAlpaca, DeepSeek-R1)  & 860.70  & 446.98 & \textcolor{orange}{412.53} & \textbf{\textcolor{blue}{327.73}} & 435.29 & \textbf{\textcolor{red}{269.06}} \\
    (Evol-CodeAlpaca, Llama)        &  699.22 & 292.19 & 310.87 & \textbf{\textcolor{blue}{279.26}} & \textbf{\textcolor{orange}{290.88}} & \textbf{\textcolor{red}{234.09}} \\
    \bottomrule
  \end{tabular*}
\end{table*}

\begin{table*}[t]
\centering
\caption{p90 latency \textnormal{(ms/token)} under different arrival rates}
\label{tab:consolidated_p90_latency}
\resizebox{\textwidth}{!}{
\begin{tabular}{cccccccc}
\toprule
\makecell{\textbf{Dataset} / \\ \textbf{Model}} & \makecell{\textbf{Arrival Rate} \\ (requests/s)} & \textbf{FCFS} & \textbf{Pointwise SJF} & \textbf{Listwise SJF} & \textbf{PARS} & \textbf{Cross-Model PARS} & \textbf{Oracle SJF} \\
\midrule
\multirow{4}{*}{\makecell{(Alpaca, DeepSeek-R1)}} 
& 5  & 1253.48 & 942.43 & 432.01 & \textbf{\textcolor{blue}{405.29}} & \textbf{\textcolor{orange}{522.88}} & \textbf{\textcolor{red}{255.56}} \\
& 10 & 1974.66 & 1420.63 & 781.94 & \textbf{\textcolor{blue}{627.68}} & \textbf{\textcolor{orange}{999.97}} & \textbf{\textcolor{red}{488.76}} \\
& 15 & 2525.96 & 1609.07 & 846.69 & \textbf{\textcolor{blue}{708.01}} & \textbf{\textcolor{orange}{1098.85}} & \textbf{\textcolor{red}{537.84}} \\
& 20 & 2791.17 & 1685.89 & 1014.15 & \textbf{\textcolor{blue}{773.33}} & \textbf{\textcolor{orange}{1199.60}} & \textbf{\textcolor{red}{576.92}} \\
\midrule
\multirow{4}{*}{\makecell{(Alpaca, Llama)}} 
& 5  & 57.01 & 78.25 & 83.92 & \textbf{\textcolor{blue}{48.90}} & \textbf{\textcolor{orange}{50.76}} & \textbf{\textcolor{red}{31.72}} \\
& 10 & 185.44 & 476.65 & 367.18 & \textbf{\textcolor{blue}{153.31}} & \textbf{\textcolor{orange}{168.37}} & \textbf{\textcolor{red}{124.86}} \\
& 15 & 757.52 & 650.62 & 584.50 & \textbf{\textcolor{blue}{236.17}} & \textbf{\textcolor{orange}{301.31}} & \textbf{\textcolor{red}{148.29}} \\
& 20 & 2169.57 & 644.10 & 631.10 & \textbf{\textcolor{blue}{270.32}} & \textbf{\textcolor{orange}{304.91}} & \textbf{\textcolor{red}{203.69}} \\
\midrule
\multirow{4}{*}{\makecell{(LMSYS-Chat-1M, DeepSeek-R1)}}
& 5  & 2259.32 & 1457.28 & 1052.87 & \textbf{\textcolor{blue}{709.26}} & \textbf{\textcolor{orange}{978.57}} & \textbf{\textcolor{red}{477.84}} \\
& 10 & 3045.16 & 1866.41 & 1296.64 & \textbf{\textcolor{blue}{1008.30}} & \textbf{\textcolor{orange}{1289.04}} & \textbf{\textcolor{red}{691.44}} \\
& 15 & 3104.24 & 1970.61 & 1488.47 & \textbf{\textcolor{blue}{1093.94}} & \textbf{\textcolor{orange}{1430.72}} & \textbf{\textcolor{red}{813.00}} \\
& 20 & 3441.00 & 2187.29 & 1586.65 & \textbf{\textcolor{blue}{1130.98}} & \textbf{\textcolor{orange}{1538.22}} & \textbf{\textcolor{red}{838.34}} \\
\midrule
\multirow{4}{*}{\makecell{(LMSYS-Chat-1M, Llama)}} 
& 5  & 1728.65 & 552.08 & 315.74 & \textbf{\textcolor{blue}{124.44}} & \textbf{\textcolor{orange}{124.49}} & \textbf{\textcolor{red}{115.90}} \\
& 10 & 5220.32 & 1121.03 & 779.95 & \textbf{\textcolor{blue}{629.98}} & \textbf{\textcolor{orange}{701.62}} & \textbf{\textcolor{red}{287.53}} \\
& 15 & 6374.70 & 1469.93 & 811.49 & \textbf{\textcolor{blue}{712.62}} & \textbf{\textcolor{orange}{767.40}} & \textbf{\textcolor{red}{310.74}} \\
& 20 & 7510.52 & 1537.07 & 867.26 & \textbf{\textcolor{blue}{784.07}} & \textbf{\textcolor{orange}{852.20}} & \textbf{\textcolor{red}{405.02}} \\
\midrule
\multirow{4}{*}{\makecell{(MATH-plus, DeepSeek-R1)}} 
& 5  & 908.07 & 108.32 & \textbf{\textcolor{orange}{92.10}} & \textbf{\textcolor{blue}{69.64 }} & 103.19 & \textbf{\textcolor{red}{57.73}} \\
& 10 & 1885.80 & 257.91 & \textbf{\textcolor{orange}{163.33}} & \textbf{\textcolor{blue}{154.46}} & 335.67 & \textbf{\textcolor{red}{108.39}} \\
& 15 & 2183.58 & 341.71 & \textbf{\textcolor{orange}{262.16}} & \textbf{\textcolor{blue}{221.75}} & 457.86 & \textbf{\textcolor{red}{160.55}} \\
& 20 & 2382.07 & 436.16 & \textbf{\textcolor{orange}{335.07}} & \textbf{\textcolor{blue}{279.02}} & 550.67 & \textbf{\textcolor{red}{204.69}} \\
\midrule
\multirow{4}{*}{\makecell{(MATH-plus, Llama)}} 
& 5  & 4453.90 & 1064.15 & 1162.96 & \textbf{\textcolor{blue}{640.16}} & \textbf{\textcolor{orange}{1050.13}} & \textbf{\textcolor{red}{84.49}} \\
& 10 & 5918.61 & 1458.51 & 1882.49 & \textbf{\textcolor{blue}{1136.75}} & \textbf{\textcolor{orange}{1335.92}} & \textbf{\textcolor{red}{117.38}} \\
& 15 & 6860.07 & 1844.42 & 2261.93 & \textbf{\textcolor{blue}{1347.30}} & \textbf{\textcolor{orange}{2064.08}} & \textbf{\textcolor{red}{131.78}} \\
& 20 & 6901.16 & 2038.44 & 2193.29 & \textbf{\textcolor{blue}{1345.71}} & \textbf{\textcolor{orange}{2043.99}} & \textbf{\textcolor{red}{168.85}} \\
\midrule
\multirow{4}{*}{\makecell{(Evol-CodeAlpaca, DeepSeek-R1)}}
& 5  & 2367.49 & 1366.32 & \textbf{\textcolor{orange}{983.09}} & \textbf{\textcolor{blue}{776.19}} & 1220.01 & \textbf{\textcolor{red}{662.57}} \\
& 10 & 2794.57 & 1672.61 & \textbf{\textcolor{orange}{1250.36}} & \textbf{\textcolor{blue}{1138.30}} & 1598.63 & \textbf{\textcolor{red}{824.24}} \\
& 15 & 3044.98 & 1707.16 & \textbf{\textcolor{orange}{1364.53}} & \textbf{\textcolor{blue}{1140.19}} & 1623.47 & \textbf{\textcolor{red}{903.28}} \\
& 20 & 3208.29 & 1915.15 & \textbf{\textcolor{orange}{1449.33}} & \textbf{\textcolor{blue}{1188.03}} & 1823.40 & \textbf{\textcolor{red}{967.25}} \\
\midrule
\multirow{4}{*}{\makecell{(Evol-CodeAlpaca, Llama)}} 
& 5  & 1971.66 & 700.83 & 672.27 & \textbf{\textcolor{blue}{485.93}} & \textbf{\textcolor{orange}{508.17}} & \textbf{\textcolor{red}{403.49}} \\
& 10 & 2776.77 & 972.18 & \textbf{\textcolor{orange}{869.86}} & \textbf{\textcolor{blue}{728.53}} & 910.77 & \textbf{\textcolor{red}{620.20}} \\
& 15 & 3137.83 & 1098.74 & 1185.43 & \textbf{\textcolor{blue}{742.42}} & \textbf{\textcolor{orange}{846.89}} & \textbf{\textcolor{red}{698.37}} \\
& 20 & 3418.86 & 1179.92 & 1422.59 & \textbf{\textcolor{blue}{813.92}} & \textbf{\textcolor{orange}{941.35}} & \textbf{\textcolor{red}{742.17}} \\
\bottomrule
\end{tabular}}
\end{table*}

\subsection{Throughput and Batching Efficiency}

\label{section:throughput}

While PARS significantly reduces per-token latency, an important concern is whether such improvements come at the cost of reduced batching efficiency. In particular, aggressive reordering may reduce batch homogeneity and hurt overall throughput. To evaluate this, we measure the completed request throughput (requests per second) under fixed-load conditions with varying request rates and burst workloads. This setup allows us to directly assess whether latency improvements are achieved without sacrificing system throughput.

Table~\ref{tab:throughput} shows the throughput comparison between FCFS and PARS. Across all settings, PARS maintains or slightly improves throughput. For example, under 20 QPS, PARS improves throughput from 2.57 to 2.87 requests per second. Similar trends are observed under both steady and bursty workloads. These results demonstrate that PARS's latency improvements do not come at the expense of batching efficiency or overall throughput. On the contrary, by grouping requests with similar expected decode lengths, PARS increases batch homogeneity, which leads to more efficient GPU utilization during decoding. 

\begin{table}[t]
\centering
\caption{Throughput (completed requests per second) under varying load conditions (Alpaca dataset, DeepSeek-R1 model).}
\label{tab:throughput}
\begin{tabular}{lccc}
\toprule
\textbf{Load} & \textbf{FCFS (req/s)} & \textbf{PARS (req/s)} \\
\midrule
Burst Mode & 2.33 & 2.51 \\
5 QPS      & 2.48 & 2.50 \\
10 QPS     & 2.70 & 2.73 \\
15 QPS     & 2.56 & 2.78 \\
20 QPS     & 2.57 & 2.87 \\
\bottomrule
\end{tabular}
\end{table}

\subsection{Impact of Starvation Prevention}
\label{section:starvation}

Figure~\ref{fig:latency_starvation_comparison} illustrates the scheduling efficiency with and without the starvation prevention mechanism. While our approach introduces a slight increase in average latency, it significantly reduces the p90 latency, particularly under high request arrival rates. This demonstrates that our method enhances tail performance and fairness with minimal impact on overall system efficiency.

\subsection{Runtime Overhead} 
\label{section:overhead}

Accurate ranking is critical to effective scheduling, but its added overhead must remain minimal to preserve responsiveness in LLM serving. We analyze the overhead introduced by PARS from two perspectives:
First, PARS continuously computes a priority score for each incoming request before it enters the waiting queue. This process runs in the background and is non-blocking, meaning it does not interfere with the LLM’s actual execution. LLMs automatically select high-priority requests for execution without waiting on the scoring process. As a result, PARS does not introduce any additional latency to request execution. The only resource cost is a lightweight GPU memory footprint of approximately 1000MB to support PARS running in the background.

Second, the computational cost of sorting and ranking is minimal. On average, PARS takes approximately 0.2 seconds to score and sort a batch of 100 prompts, while processing 100 actual LLM requests takes around 250 seconds. This corresponds to just 0.08\% of the total serving time. These results show that even under high-load conditions, PARS can swiftly assign priority scores to incoming requests, ensuring responsiveness is consistently maintained.

\relax

\section{Discussion}

In this section, we discuss how PARS generalizes to different workload characteristics and deployment settings beyond those explicitly evaluated in our experiments.

\subsection{Behavior under Input-Dominated Workloads}

PARS is primarily designed to mitigate head-of-line blocking caused by heterogeneity in request execution times, especially when requests differ substantially in output lengths during decoding. However, in workloads such as summarization or retrieval augmented generation, requests are often dominated by long input prompts and relatively short outputs. In these cases, the total execution time is largely determined by the prefill phase, which depends on the input length and is known prior to execution. 

As a result, scheduling decisions do not require prediction, since requests can be directly ordered using their input lengths to achieve exact SJF scheduling. Under such conditions, PARS effectively becomes a deterministic scheduler that leverages known input lengths, eliminating the need for predictive ranking. Therefore, PARS remains efficient and does not introduce unnecessary overhead in input dominated workloads.

\subsection{Extension to Multi-node and Disaggregated Serving}

PARS is designed as a lightweight scheduling layer that operates independently of the underlying execution engine, making it naturally compatible with both distributed multi-node deployments and prefill/decode-disaggregated serving architectures. In these systems, request routing is typically handled by a centralized proxy or dispatcher that assigns incoming requests to worker nodes. PARS can be directly integrated into this proxy layer by embedding the predictor and replacing standard dispatching with predictor-guided approximate SJF scheduling.

Because PARS only modifies request ordering without altering the execution pipeline, it can be applied across different serving architectures with minimal integration effort. This makes PARS a practical and extensible solution for real-world LLM serving, including large-scale distributed and disaggregated deployments.

\section{Conclusion}
In this paper, we presented PARS, a prompt-aware LLM scheduler that leverages pairwise learning-to-rank to approximate shortest-job-first scheduling. 
We integrated PARS into the state-of-the-art LLM serving system vLLM, and conducted extensive evaluations across diverse scenarios. Results show that PARS consistently outperforms existing methods in terms of average and tail latencies, while incurring minimal ovrehead.  Moreover, PARS generalizes effectively across LLM architectures and datasets, demonstrating its robustness and practicality for low-latency LLM serving. 

The need for scheduling strategies that mitigate HOL blocking will only intensify as reasoning-capable LLMs become more prevalent.
This work shows that a lightweight, prompt-aware ranking mechanism like PARS is sufficient to capture relative execution cost for effective scheduling, providing a practical, deployable alternative to heavyweight or model-specific approaches for real-world LLM serving.

While our experiments demonstrate the clear benefits of using PARS for LLM serving, the magnitude of the gains \emph{depends heavily on task length variability}. When task durations are nearly identical, the room for optimization is minimal, and PARS yields only marginal improvements over FCFS. In contrast, PARS is most effective when task execution times differ significantly, as prioritizing shorter tasks helps mitigate HOL blocking and substantially reduces latency. We also acknowledge that for certain workloads where prefill latency is non-negligible (e.g., summarization with long inputs), a simple SJF policy based only on output length may be insufficient; we leave this direction for future exploration, along with extending PARS to multi-modal environments, addressing challenges such as inter-node communication, load balancing, and distributed resource management.

The software tool ($\sim$1,000 lines of code), along with the associated datasets, is available as open-source code on GitHub at \href{https://github.com/SPEAR-UIC/PARS}{https://github.com/SPEAR-UIC/PARS}.

\section*{Acknowledgement}
This work was supported in part by the U.S. National Science Foundation under Grants OAC-2402901 and CCF-2515009 and by the U.S. Department of Energy under Contract DE-SC0024271.  

\bibliographystyle{unsrt}
\bibliography{ref}

\end{document}